\pdfoutput=1
\documentclass[11pt]{article}
\usepackage{latex/acl}

\usepackage{times}
\usepackage{latexsym}
\usepackage[T1]{fontenc}
\usepackage[utf8]{inputenc}
\usepackage{microtype}

\usepackage{inconsolata}
\usepackage{graphicx}
\usepackage{enumitem}
\usepackage{algorithm}
\usepackage{algpseudocode}
\usepackage{booktabs}
\usepackage{multirow}
\usepackage{makecell}
\usepackage{amssymb} 
\usepackage{pifont} 
\usepackage{colortbl}
\usepackage{subfigure}
\usepackage{amsmath}
\usepackage{listings}
\usepackage[normalem]{ulem}

\usepackage{natbib}

\newcommand{\eg}{\textit{e.g.~}}
\newcommand{\etc}{\textit{etc.~}}
\newcommand{\ie}{\textit{i.e.~}}

\newcommand{\grayline}{\rowcolor[gray]{.90}}

\definecolor{codegreen}{rgb}{0,0.6,0}
\definecolor{codegray}{rgb}{0.5,0.5,0.5}
\definecolor{codepink}{RGB}{252, 142, 172}
\definecolor{codepurple}{rgb}{0.58,0,0.82}
\definecolor{backcolour}{RGB}{245,245,245}
\lstdefinestyle{mystyle}{
    backgroundcolor=\color{backcolour},   
    commentstyle=\color{magenta},
    keywordstyle=\color{blue},
    numberstyle=\tiny\color{codegray},
    stringstyle=\color{codepurple},
    basicstyle=\fontfamily{\ttdefault}\footnotesize,
    breakatwhitespace=false,         
    breaklines=true,                 
    keepspaces=true,    
    frame=single,
    numbersep=5pt,                  
    showspaces=false,                
    showstringspaces=false,
    showtabs=false,                  
    tabsize=2,
    classoffset=1, 
    keywordstyle=\color{violet},
    classoffset=0,
}
\title{FragRel: Exploiting Fragment-level Relations in the External Memory of Large Language Models}

\author{
\centerline{\makecell{Xihang Yue, Linchao Zhu, Yi Yang$^\dag$}}\\
\centerline{ ReLER, CCAI, Zhejiang University }\\
\centerline{\tt{\{xihang,zhulinchao,yangyics\}@zju.edu.cn}} \\
\centerline{ $^\dag$ Corresponding author }
}

\begin{document}
\maketitle

\begin{abstract}

To process contexts with unlimited length using Large Language Models (LLMs), 
recent studies explore hierarchically managing the long text.
Only several text fragments are taken from the external memory and passed into the temporary working memory, i.e., LLM's context window.
However, existing approaches isolatedly handle the text fragments without considering their structural connections, thereby suffering limited capability on texts with intensive inter-relations, e.g., coherent stories and code repositories.
This work attempts to resolve this by exploiting the fragment-level relations in external memory.
First, we formulate the fragment-level relations and present several instantiations for different text types.
Next, we introduce a relation-aware fragment assessment criteria upon previous independent fragment assessment.
Finally, we present the fragment-connected Hierarchical Memory based LLM.
We validate the benefits of involving these relations on long story understanding, repository-level code generation, and long-term chatting.
\end{abstract}

\section{Introduction}
The limited context window length constrains applications of Large Language Models (LLMs) in some practical scenarios, such as answering questions based on complete books or movie scripts~\citep{kovcisky2018narrativeqa}, writing codes within complete Github repositories~\citep{zhang2023repocoder}, \etc
To resolve this problem, some works~\citep{ding2023longnet,han2023lm,xiao2023efficient} attempt to expand the context length of classical LLM inference framework via continual training or sparse attention mechanism.
However, existing approaches are either limited to a finite expansion length~\citep{packer2023memgpt,schuurmans2023memory}, or prone to performance degradation, especially when dealing with very long contexts~\citep{liu2023lost}.

Recent studies~\citep{packer2023memgpt,wang2023augmenting,ram2023context} explore to hierarchically process the long text.
Each time 
only partial fragments of long text
are retrieved from external memory and fed into LLM's context window, a.k.a, temporary working memory, thereby eliminating the context length constraint and alleviating the inferior influence of substantial irrelevant content. 
However, current external memory managers simply split the complete long context into independent fragments, assessing their isolated importance during retrieval.
This constrains the inference capability of Hierarchical Memory based LLMs, particularly in scenarios (\eg understanding coherent story or code repository) where there are intensive associations across fragments of long text.

\begin{figure*}[htbp]
\begin{center}
\includegraphics[width=\linewidth]{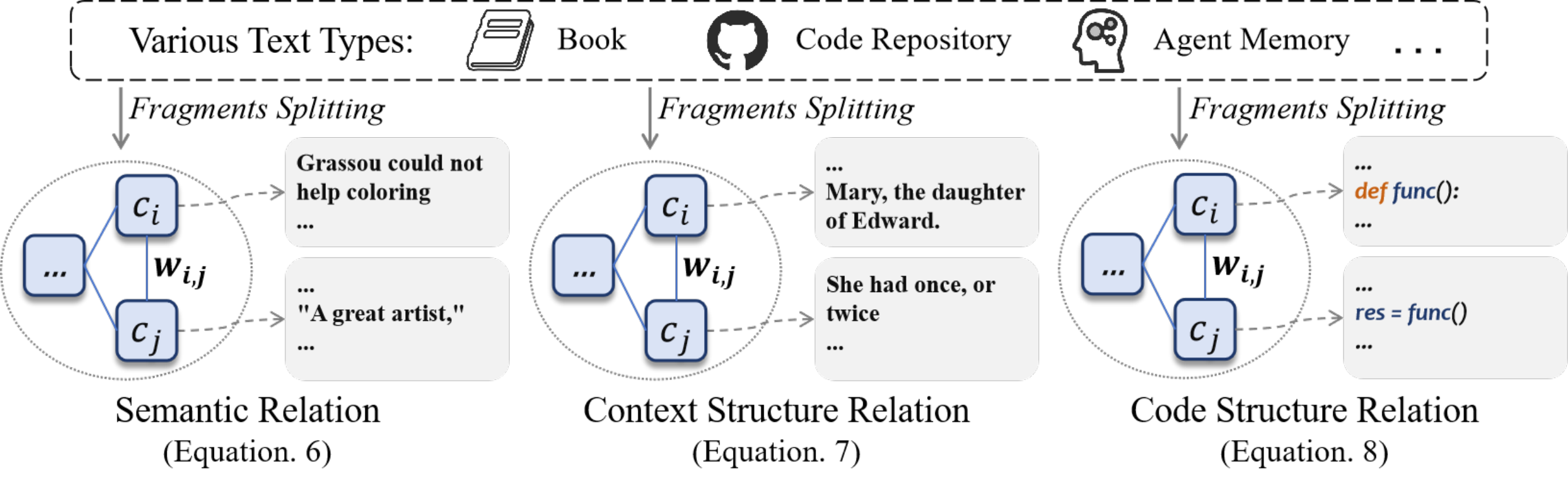}
\caption{The instantiated fragment-level relations in various text types.}
\label{fig:hiera_llm}
\end{center}
\end{figure*}

To address this issue, we propose to integrate these fragment-level relations into the external memory management by introducing a relation-aware fragments assessment score during retrieval.
First, we formulate the relations between two fragments as a real number, with higher values corresponding to stronger relation strength.
The calculation of relation quantity could have different instantiations for different context types (\eg narrative stories, code repositories, or historical dialog) and different relation types (\eg semantic relations or structural relations).
Next, based on the isolated relevance scores used in past external memory retrievers, we further calculate every fragment's environmental relevance score, which is defined as a normalized relation-weighted summation of other fragments' independent scores.
During retrieval, the combination of independent score and environmental score is employed for assessing every fragment's importance.
An adjustable coefficient is introduced to control the influence of environmental score.
Finally, in the same as previous works~\citep{packer2023memgpt,ram2023context}, we concatenate the retrieved content and the requested instruction as LLM inference input.

Extensive experiments validate the benefits of incorporating these fragment-level relations during retrieval.
The experiments encompass a variety of base LLMs (Llama2~\citep{touvron2023llama}, ChatGPT, ChatGLM~\citep{du2022glm}, \etc), different temporary context lengths (1K, 4K, 8K, 20K, \etc), and multiple long context scenarios (Long Story Understanding~\citep{kovcisky2018narrativeqa}, Repository-level Code Completion~\citep{zhang2023repocoder}, and Long-term Chatting with Human~\citep{lu2023memochat}).

\section{Related Work}

\paragraph{Temporarily Long Context Processing.}
One line of works explores how to process long context under the typical LLM inference framework, in which the complete context is directly stored in the LLM context window (\ie temporary memory).
Among them, a series of  works~\citep{dai2019transformer,ding2023longnet,han2023lm,xiao2023efficient,xiong2023effective} study extending the context length of Transformer-based models via efficient attention mechanism and recurrent inference strategic.
In addition, some works~\citep{li2023unlocking,jiang2023longllmlingua} investigate how to compress the length of long text content to mitigate the impact of excessive irrelevant text.
Although the long text processing capability of the large language models can be enhanced by expanding the context window or compressing the context content,
the context length that can be handled remains limited.

\paragraph{External Memory augmented LLMs.}
Another line of works introduce an additional external memory, forming a hierarchical memory based inference framework, thereby processing context of any length.
In this framework, 
only partial content fragments are retrieved from external memory for knowledge updating~\citep{wu2022memorizing,wang2023augmenting} or answering knowledge-intensive questions~\citep{lewis2020retrieval,guu2020retrieval,borgeaud2022improving,lan2023copy,wang2023instructretro}. 
The most popular retrieval method is first calculating the text embedding similarity for isolated fragments of external context using pre-trained embedding models~\citep{su2022one,zhang2023retrieve}, and then  retrieving the text fragments with higher similarity to the requested question or current temporary context.

Benefiting from the zero-shot generalization capability of LLMs, the retrieved fragments can be directly concatenated with instructions as model input~\citep{ram2023context}, eliminating the need for additional training.
This further facilitates more flexible external memory augmented LLM inference frameworks.
Some studies explore the collaborative use of retrieval and generation~\citep{gao2022precise,yan2024corrective}, as well as further multi-round retrieval-generation interleaving framework~\citep{trivedi2022interleaving,jiang2023active,asai2023self,shao2023enhancing,feng2023retrieval}.
\citet{saad2023pdftriage} utilizes explicit prompts about the structure of external context for enhanced retrieval.
Additionally, MemGPT~\citep{packer2023memgpt} automatically reads and writes the external memory, enabling more flexible external memory reasoning.

\section{Methodology}

\begin{figure*}[ht]
\begin{center}
\includegraphics[width=1.0\linewidth]{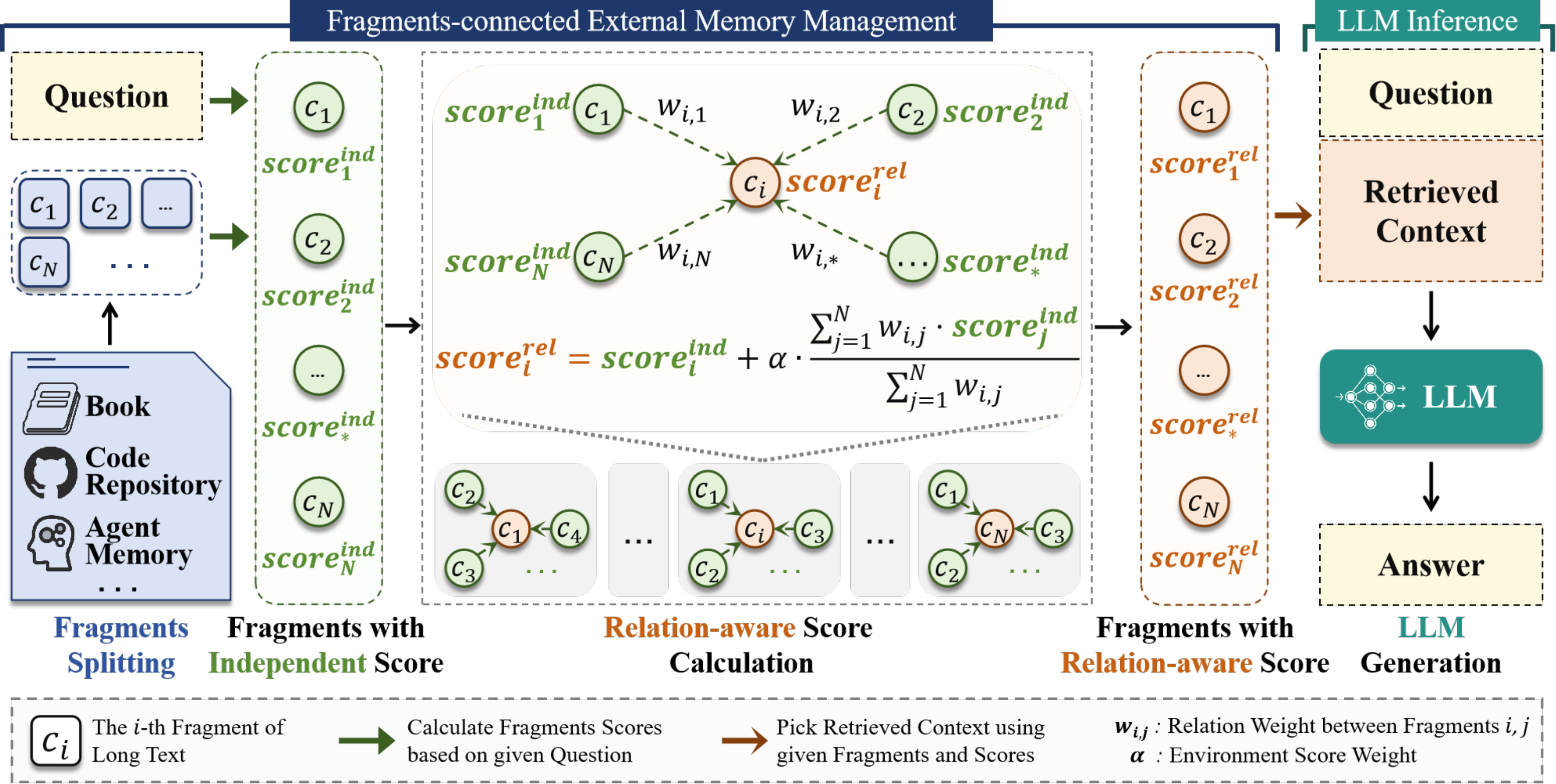}
\caption{Overall framework of Fragment-connected Hierarchical Memory based LLM.
First, the long context (complete book, code repository, or agent memory) is split into a lot of fragments $c_*$. 
Then we calculate the independent similarity score $score^{ind}_*$ between every fragment and the user question. 
Next, for every fragment, its relation-aware score $score^{rel}_*$ is calculated as the combination of the independent score and the normalized relation-weighted summation of other fragments' independent score.
Finally, the fragments with top $K$ relation-aware scores are retrieved for LLM inference.
}
 \label{fig:model_structure}
\end{center}
\end{figure*}

\subsection{Preliminary}

\subsubsection{Temporary Memory based LLM}

\label{sec:tempmem-llm}
\paragraph{Formulation.}

Traditional language models receive input $x$ and generate the output $y$ by:
\begin{equation}
\label{equ:tempmem_llm_singleinput}
y = \text{LLM}(x).
\end{equation}
The input $x$ is straightly passed into the context window of LLM.
Notably, the instructed LLMs~\citep{ouyang2022training} could take various user instructions $x$  and produce the corresponding responses $y$.
Inspired by human cognition, the LLM context window could be viewed as the \emph{working memory} which temporarily stores information~\citep{packer2023memgpt,li2022large}.
For clarity, we refer to this traditional LLM reasoning framework shown in Equation.~\ref{equ:tempmem_llm_singleinput} as \emph{Temporary Memory based LLM} (TempMem-LLM).

\paragraph{Long Text Processing.}

In many practical tasks, the input content includes not only user instructions $x$, but also an additional long context $\mathcal{T}$, such as answering questions based on complete storybooks or movie scripts~\citep{kovcisky2018narrativeqa}, writing codes in long Github repository~\citep{zhang2023repocoder}, and constructing agents capable of engaging in long-term conversations~\citep{lu2023memochat,zhong2023memorybank}.
In these scenarios, TempMem-LLM simply concatenates and processes the long context $\mathcal{T}$ and instruction $x$ in its working memory:
\begin{equation}
\label{equ:formulation_templlm_longtext}
y=\text{LLM}(x \oplus \mathcal{T}), 
\end{equation}
where $\oplus$ represents the concatenation operation.

When the text length of $\mathcal{T}$ exceeds the context window limitations of LLM, we could cut its additional content of $\mathcal{T}$ for executing inference.
In next section, we present another framework for effectively processing the lengthy context $\mathcal{T}$ via storing it in external memory and retrieving relevant fragments for inference every time.

\subsubsection{Hierarchical Memory based LLM}
\label{sec:hieramem-llm}
Unlike TempMem-LLM which handles all context in its temporary working memory, the Hierarchical Memory based LLM (HieraMem-LLM) integrates an additional non-parametric external memory for managing the long context $\mathcal{T}$. 

\paragraph{Formulation.}
Instead of directly being concatenated with the user instruction $x$, the context $\mathcal{T}$ is independently processed in HieraMem-LLM.
Therefore, it
consists of two decoupled modules, i.e., the external memory management module processing the long context $\mathcal{T}$, as well as the LLM forward inference module containing the temporary working memory. Formally, we have: 
\begin{equation}
\begin{aligned}
\mathcal{T}^{ret} &= 
\text{Mem-MGR}
(x, \mathcal{T}), \\
y &= \text{LLM}(x \oplus \mathcal{T}^{ret}).
\end{aligned}
\end{equation}
Mem-MGR is the external memory manager, which takes the  requested instruction $x$ as input and returns  relevant fragments
from $\mathcal{T}$.

\paragraph{External Memory Manager.} 
The typical external memory manager consists of three steps, i.e., fragment splitting, 
independent score calculation,
and fragment selection.

\noindent1. \emph{Fragment Splitting}. 
It splits the long text $\mathcal{T}$ into $N$ short fragments $c_*$. 

\noindent2. \emph{Independent Score Calculation}. 
It calculates the independent score $s_i^{ind}$ (or $score_i^{ind}$) for every fragment $c_i$ based on user input $x$, i.e., \begin{equation}
\label{equ:independent_score_calculation}
s_i^{ind} = \text{Similarity}(x, c_i).
\end{equation}
The similarity function 
is often instantiated as the cosine similarity between embedding vectors of $x$ and $c_i$, which could be calculated using pre-trained text embedding models, \eg the Contriever model~\citep{izacard2021contriever}.

\noindent3. \emph{Retrieved Context Picking}. With fragment scores $s_i^{ind}$, we select related fragments with top $K$ scores and feed them into the temporary memory of LLM.

Although HieraMem-LLM effectively manages the long context $\mathcal{T}$ by utilizing external memory, the context is decomposed into unrelated segments, disrupting the structural integrity of the context.

\paragraph{TempMem \textit{vs.} HieraMem.}
TempMem-LLM straightly handles the complete long text $\mathcal{T}$ in its working memory, integrating token-level correlations during inference.
The simple concatenation approach in Equation~\ref{equ:formulation_templlm_longtext} suffers the following issues:
(1)~When $\mathcal{T}$ exceeds the model's context length constraint, the LLM is unable to do prediction.
(2)~The information irrelevant to instruction $x$ can interfere with the model's processing, leading to performance decline~\citep{liu2023lost}.
(3)~Reprocessing the lengthy context $\mathcal{T}$ each time consumes excessive  computational resources.
To address the issues of TempMem-LLM, 
HieraMem-LLM incorporates the external memory to manage the prolix context $\mathcal{T}$. Only a few related fragments are extracted for LLM inference.

However, existing external memory managers select fragments based on only  isolated fragment content, overlooking intensive relations between fragments.
While in TempMem-LLM, the long-term correlations could be employed via the attention mechanism over the complete text, enabling comprehensive context modeling.

\subsection{Fragment Relations}

\subsubsection{Definition}
\label{sec:rel_fragments_definition}

We formulate the relations between every pair of fragments $(c_i,c_j)$ as follows:
\begin{equation}
\label{equ:fragments_relation_definition}
w_{i,j}=\mathcal{F}^{rel}(c_i,c_j), \,1 \le i,j \le N,
\end{equation}
where $w_{i,j}$ is a real numbers measuring the relation strength between fragment $c_i$ and fragment $c_j$.
The larger value of $w_{i,j}$ indicates the stronger correlation between $c_i$ and $c_j$.
Next, we present several specific implementations for $\mathcal{F}^{rel}$.

\subsubsection{Fragment Relation Instantiations}
\label{sec:rel_fragments_extraction}

This section introduces several instantiations of fragment-level relations and discusses the importance of considering these relations.

\paragraph{Semantic Relation.}
Semantic association is the most common type of connection between text segments.
The semantic correlation between text fragments can be measured by the cosine similarity between the latent embeddings of fragments,
\begin{equation}
\label{equ:relation_instantiation_semantic}
\mathcal{F}^{rel}(c_i, c_j) = 1 - { e_{i} \cdot e_{j}  \over  \| e_{i} \| \| e_{j} \| },
\end{equation}
where $e_{i}$ and $e_{j}$ represent the latent embeddings of fragments $c_i$ and $c_j$ respectively.
They could be obtained with the pre-trained embedding models, \eg the Contriever model~\citep{izacard2021contriever}.

\paragraph{Context Structure Relation.}
In consecutive books or long-term dialog, there are significant content  correlations between the contiguous preceding and following fragments. The fragments with  closer positions in the context have stronger relevance.
This contextual relationship strength can be defined as:
\begin{equation}
\label{equ:relation_instantiation_context_structure}
\mathcal{F}^{rel}(c_i, c_j) = w_{rel}^{ | loc_i - loc_j | }, \, w_{rel} \in [0,1],
\end{equation}
where $loc_i$ refers the absolute position of fragment $c_i$ in the external context $\mathcal{T}$.
$w_{rel}$ can be adjusted to control the relation strength between fragments.
When $w_{rel}=0$, it represents there is no relation between fragments, and our method degrades to previous fragment-independent external memory.

\paragraph{Code Structure Relation.}
Compared to natural language, code repository fragments have more complex interrelations.
We construct the structure graph $\mathcal{G}$ for a complete code  repository.
The code graph $\mathcal{G}$ consists of all code parsing nodes (including function definition, function body, assignment expression, \etc).
The code parsing nodes are connected by edges based on the parsing tree, function calling relation, and the files directory structure. 
The edge weights take values in $[0,1]$ and are set based on the edge types.
In section.~\ref{sec:apen_relation_code} we present more details about the construction of the code graph $\mathcal{G}$.
Based on the code graph $\mathcal{G}$, we formulate the code structure relation as follows:
\begin{equation}
\begin{gathered}
\label{equ:relation_instantiation_code_structure}
\mathcal{F}^{rel}(c_i, c_j) = { { \sum_{k=1}^{N^{G}_{c_i}}  \sum_{l=1}^{N^{G}_{c_j}} K^{c_i,c_j}_{k,l} \cdot \text{Dis}( g^{c_i}_k, g^{c_j}_l ) } \over { \sum_{k=1}^{N^{G}_{c_i}}  \sum_{l=1}^{N^{G}_{c_j}} K^{c_i,c_j}_{k,l} } }, \\
K^{c_i,c_j}_{k,l} = len^{c_i}_k \cdot len^{c_j}_{l},
\end{gathered}
\end{equation}
where 
$N^{G}_{c_i}$ represents the number of non-overlapping graph nodes in the $i$-th repository fragment,
$g^{c_i}_k$ is the $k$-th graph node in the $i$-th repository fragment,
$len_k^{c_i}$ represents the text length of the $k$-th paring node in fragment $c_i$.
$\text{Dis}(g_k^{c_i},g_k^{c_j})$ is the shortest path distance between node $g_k^{c_i}$ and $g_k^{c_j}$ on the code graph $\mathcal{G}$.
Section.~\ref{sec:apen_relation_code} shows more details on the calculation of $\text{Dis}(g_k^{c_i},g_k^{c_j})$.

\paragraph{More Relations.}
More relations could be designed for specific text properties and practical needs.
For example, we can gauge the correlation strength between academic papers based on citation relationships and author associations.

\paragraph{Importance of Fragments Relations.}
The fragments-level relations are significant for:

\noindent \textbf{1.} Ubiquitous existence.
These fragments-level relations ubiquitously appear in a variety of long texts.
For example, in narrative books or movie scripts, the storyline progresses coherently from beginning to end, with each fragment $c_*$ intricately connected to its preceding and following fragments $c_*$.
In code repository, structural correlations exist among different code lines, and function calling or variable passing relationships exist between different code files and functions. 
Therefore, the content of different code fragments $c_*$ are densely related.

\noindent \textbf{2.} Assisting long text understanding.
Unlike TempMem-LLM latently utilizes long-term relations, 
we posit that involving explicit inter-relations plays its role in external memory management by forming a more comprehensive assessment criteria for fragment selection.
Specifically, when neglecting the fragments relations, the external memory retriever greedily supposes \uline{only the fragments with high \emph{direct similarity} to the input $x$ is helpful (Previous)}.
The consideration of fragment relations allows a more comprehensive fragments assessment criteria:
\ie \uline{the fragments with both higher \emph{direct similarity} and \emph{contextual similarity} to $x$ is important (Ours)}.
The contextual similarity of fragment $c_i$ refers to the similarity between the $c_i$'s related fragments and the input $x$.
The new assessment criteria retrieve not only directly similar fragments but also take account of: 
(a) fragments within a relevant environment, 
(b) contextual fragments of relevant fragments, which could help understand the directly relevant fragment,
(c) indirectly relevant fragments.

\subsection{Fragment-connected Memory Retrieval }
\label{sec:relation_incorporation}

This section introduces the integration of fragment relations by calculating the relation-aware assessment scores, and presents the overall framework of Fragment-connected HieraMem-LLM.

\subsubsection{Relation-aware Fragment Assessment}

Different from vanilla retriever which considers the independent importance of every fragment using independent score $s_i^{ind}$,  
we instead propose to calculate a \emph{Relation-aware Score} for more comprehensively considering the importance of every fragment.

\paragraph{Definition.} For the $i$-th fragment, the relation-aware score is composed of two parts: its independent score $s_i^{ind}$ and its \emph{environment score} $s_i^{env}$.
The independent score measures its direct relevance degree with question $x$, defined in Equation~\ref{equ:independent_score_calculation}.
The environment score $s_i^{env}$ assesses
its related fragments' relevance with question $x$.
We formulate $s_i^{env}$ (or $score_i^{env}$) as the normalized weighted summation over independent scores $s_j^{env}$ of related fragments:
\begin{equation}
\label{equ:environment_score_calculation}
    s_i^{env} = { \sum_{j} w_{i,j} \cdot s_j^{ind} \over \sum_{j} w_{i,j} },
\end{equation}
where $w_{i,j}$ is the fragment relation defined in Equation~\ref{equ:fragments_relation_definition}.
The normalization operation is introduced to ensure the consistent numerical scale of $s_i^{env}$ with $s_i^{ind}$.

Combining $s_i^{ind}$ and $s_i^{env}$, we define \emph{Relation-aware Score} $s_i^{rel}$ (or $score_i^{rel}$) of the $i$-th fragment as follows:
\begin{equation}
\label{equ:relation_aware_score_calculation}
s_i^{rel} = s_i^{ind} + \alpha \cdot s_i^{env},
\end{equation}
where $\alpha$ is an adjustable coefficient, employed to control the influence of environment score.

\paragraph{Relation Distance and Complexity.}
The utilization of explicit fragment-level relations shown in Equation.~\ref{equ:environment_score_calculation} is irrelevant with fragments distance, while TempMem-LLM extracts relations within ranges limited by the context window length.
Additionally, some complex relations (\eg code structure relations) are challenged to automatically learn while they could be employed explicitly.

\subsubsection{Fragment-connected HieraMem-LLM}

Based on the proposed {relation-aware score}, we introduce the overall framework of \emph{Fragment-connected HieraMem-LLM}, shown in Figure.~\ref{fig:model_structure}.

We first split the long text $\mathcal{T}$ into fragments and acquire their independent scores. 
Next, considering the extracted relations, we calculate the relation-aware score $s_i^{rel}$ for every fragment using Equation~\ref{equ:relation_aware_score_calculation}.
Based on these fragments along with relation-aware scores, we select relevant fragments as retrieved context:
\begin{equation}
\label{equ:pick_relation_aware_retrived_context}
\begin{gathered}
\mathcal{T}_{ret}^{rel}=c_{r_1} \oplus c_{r_2} \oplus ... \oplus c_{r_K},\\
r_1,r_2,...,r_K=arg\,TopK_{i}\,\,\, s_i^{rel},
\end{gathered}
\end{equation}
where $\oplus$ represents operation of concatenating two text fragments, $r_*$ is the indexes of retrieved fragments.
The final response is generated $y$ with LLM as follows:
\begin{equation}
\label{equ:LLM_generation_with_relation_aware_retrived_context}
\hat{y}=LLM(x, \mathcal{T}_{ret}^{rel}).
\end{equation}

\subsubsection{Discussion}

This section provides an intuitive explanation of the influence of \emph{contextual similarity} (or named environmental similarity score $s_ i^{env}$).

\noindent \textbf{1.} The fragments with high \emph{contextual similarity} are mostly beneficial for LLM inference.
Suppose fragment $c_ i$ has a high \emph{contextual similarity} $s_ i^{env}$, which refers to that $c_ i$ is situated around a "high score fragment" or within a "high score environment".
The "high score fragment" is the fragment with a large direct similarity to the input $x$. The "high score environment" is a lot of related fragments with relatively large direct similarity to the input $x$.
Therefore, the fragment $c_ i$ with high \emph{contextual similarity} could always help understand the "high score fragment" or "high score environment".

\noindent \textbf{2.} For LLMs, precisely understanding the "high score fragment" or "high score environment" is particularly important to generate correct results.
The "high score fragment" or "high score environment" is the fragment with large direct similarity to the input $x$.
If they contain beneficial information for LLM inference, the precise understanding enables LLMs to exploit this information.
If they are useless for LLM inference, the precise understanding prevents LLMs from being disturbed by them.

\noindent \textbf{3.} Additionally, the fragments with high \emph{contextual similarity} potentially contain important information for LLM inference, although they may not be directly similar to current input $x$.
Some important information is not contained in the "high score fragment" with large direct similarity to the input $x$, but situated in its surrounding fragments.
These surrounding fragments could be retrieved through a high \emph{contextual similarity}.
This phenomenon is particularly prevalent in code repository fragments retrieval, where some important information (eg. function parameters, function return type) is contained in the indirectly relevant fragments containing the "function definition".

Therefore, the performance is enhanced through retrieving the fragments with both higher direct similarity and \emph{contextual similarity} to the input $x$.

\section{Experiment}
We evaluate the proposed \emph{Fragment-connected HieraMem-LLM} on three long text understanding tasks: long story understanding (Section.~\ref{sec:exp_bookqa}), repository-level code generation (Section.~\ref{sec:exp_codegen}), and memory-enhanced chatting agent (Section.~\ref{sec:exp_chatbot}).

\subsection{Long Story Understanding}
\label{sec:exp_bookqa}

\begin{figure*}[htbp]
\centering
\includegraphics[width=1.00\textwidth]{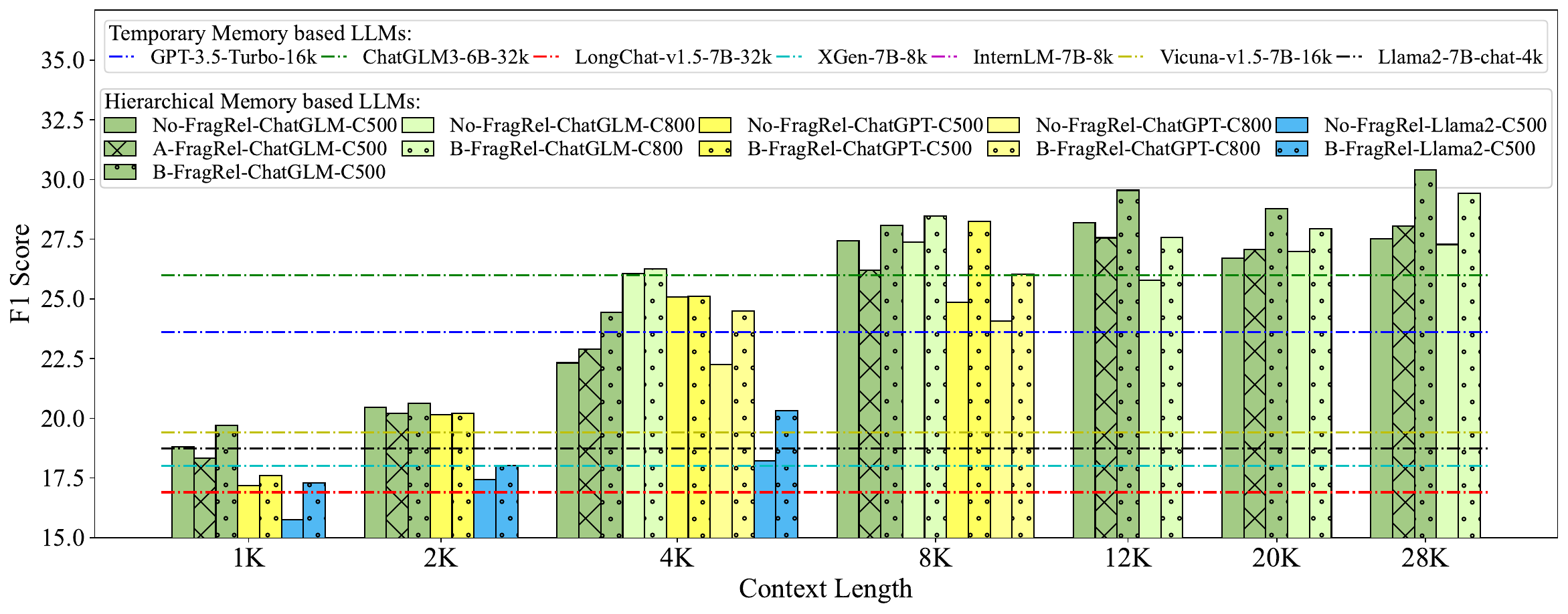}
\caption{
Performance comparison on NarrativeQA~\citep{kovcisky2018narrativeqa}.
The horizontal lines and columnas represent Temporary Memory based LLMs and Hierarchical Memory based LLMs respectively.
The Hierarchical Memory based LLMs contain 3 categories: no relation incorporation, semantic relation incorporation and context structure relation incorporation, denoted as "No-FragRel-*", "A-FragRel-*", and "B-FragRel-*" respectively.
}
\label{fig:expres_narra}
\vskip -0.2in
\end{figure*}

\subsubsection{Setup}

\paragraph{Dataset.}

NarrativeQA~\citep{kovcisky2018narrativeqa} is a challenging story comprehension benchmark, consisting of human-written question-answer paris based on long (average 18K words) story books or movie scrips.
LLMs are required to understand the long-term relations in the lengthy stories to answer these questions.
We employ the 200 extracted testing samples in LongBench~\citep{bai2023longbench}.

\paragraph{Metrics.}
Following LongBench~\citep{bai2023longbench}, we assess the generated response with the F1 Score, a widely used metric in question-answering tasks.
\paragraph{Baselines.}
We classify baselines into 2 categories: Temporary Memory based LLMs (TempMem LLMs) and Hierarchical Memory based LLMs (HieraMem LLMs).
\textbf{(a)} TempMem-LLMs: 
We directly compare the results shown in LongBench~\citep{bai2023longbench}, covering LLMs with extensive parameter amount and context length (including the context length expanded LLMs).
When the input text exceeds LLM's context length, the content in the middle position of the text is truncated.
\textbf{(b)} HieraMem-LLMs: 
We experiment with different context limitations (1K, 2K, ..., 20K, 28K words), and different fragments lengths (500 and 800 words per fragment).
The embeddings are calculated with Contriever~\citep{izacard2021contriever}, a popular pre-trained model for text retrieval.
The base LLMs includes Llama2-7B-4K~\citep{touvron2023llama}, ChatGLM3-6B-32K~\citep{du2022glm} and GPT-3.5-Turbo-16K.

\paragraph{Relation Integration Details.}

We calculate the fragment relations using semantic relation (Equation.~\ref{equ:relation_instantiation_semantic}) and context structure relation (Equation.~\ref{equ:relation_instantiation_context_structure}), denoted as "A-FragRel" and "B-FragRel" respectively.
Except for specific statements, we set $w_{rel}=0.3$ and $\alpha=0.5$.

\subsubsection{Result}

\begin{figure*}[htbp]
\centering
\subfigure[Varied $w^{rel}$]{\includegraphics[width=0.49\textwidth]{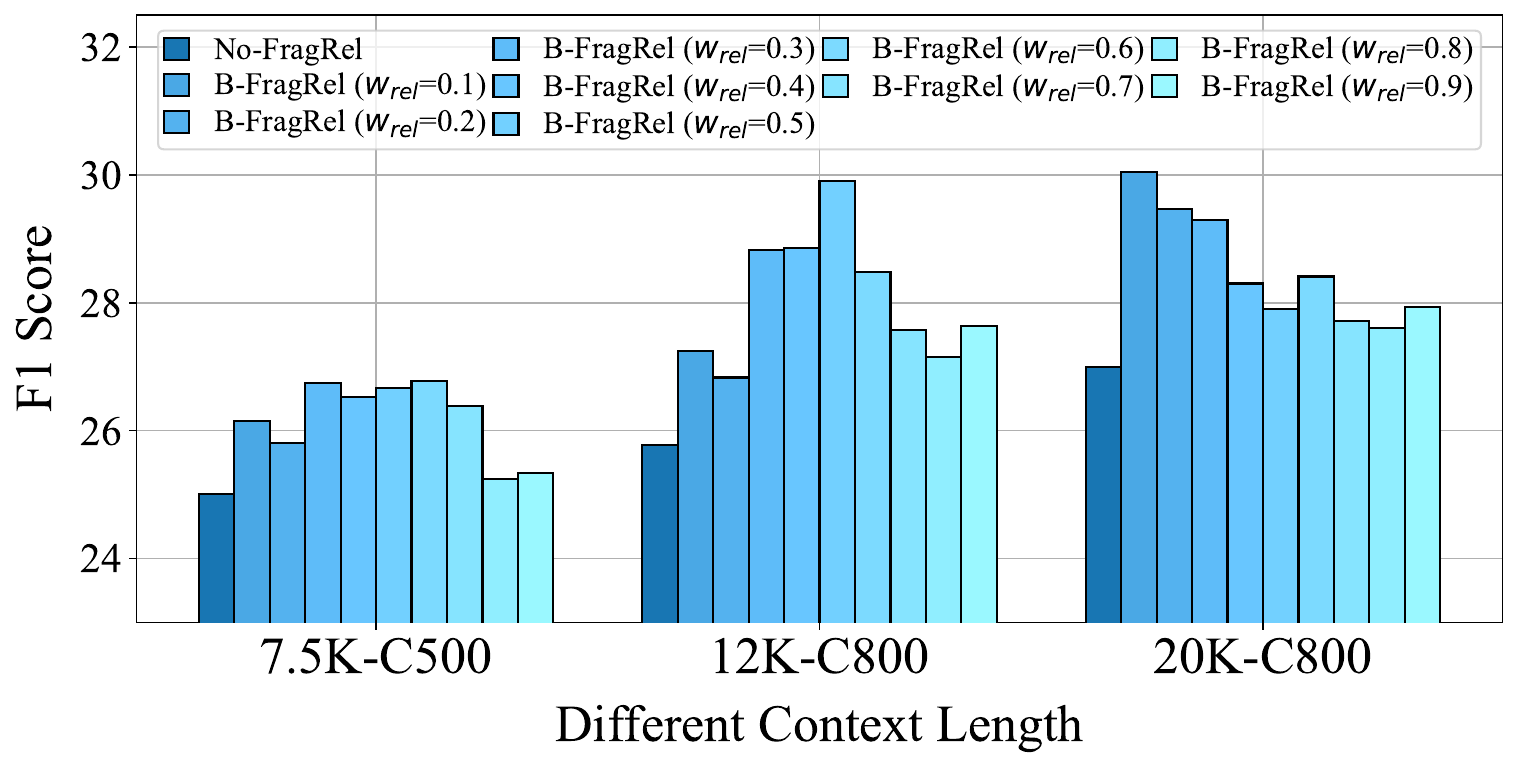}\label{fig:expres_ablation_ew}}
\subfigure[Varied $\alpha$]{\includegraphics[width=0.49\textwidth]{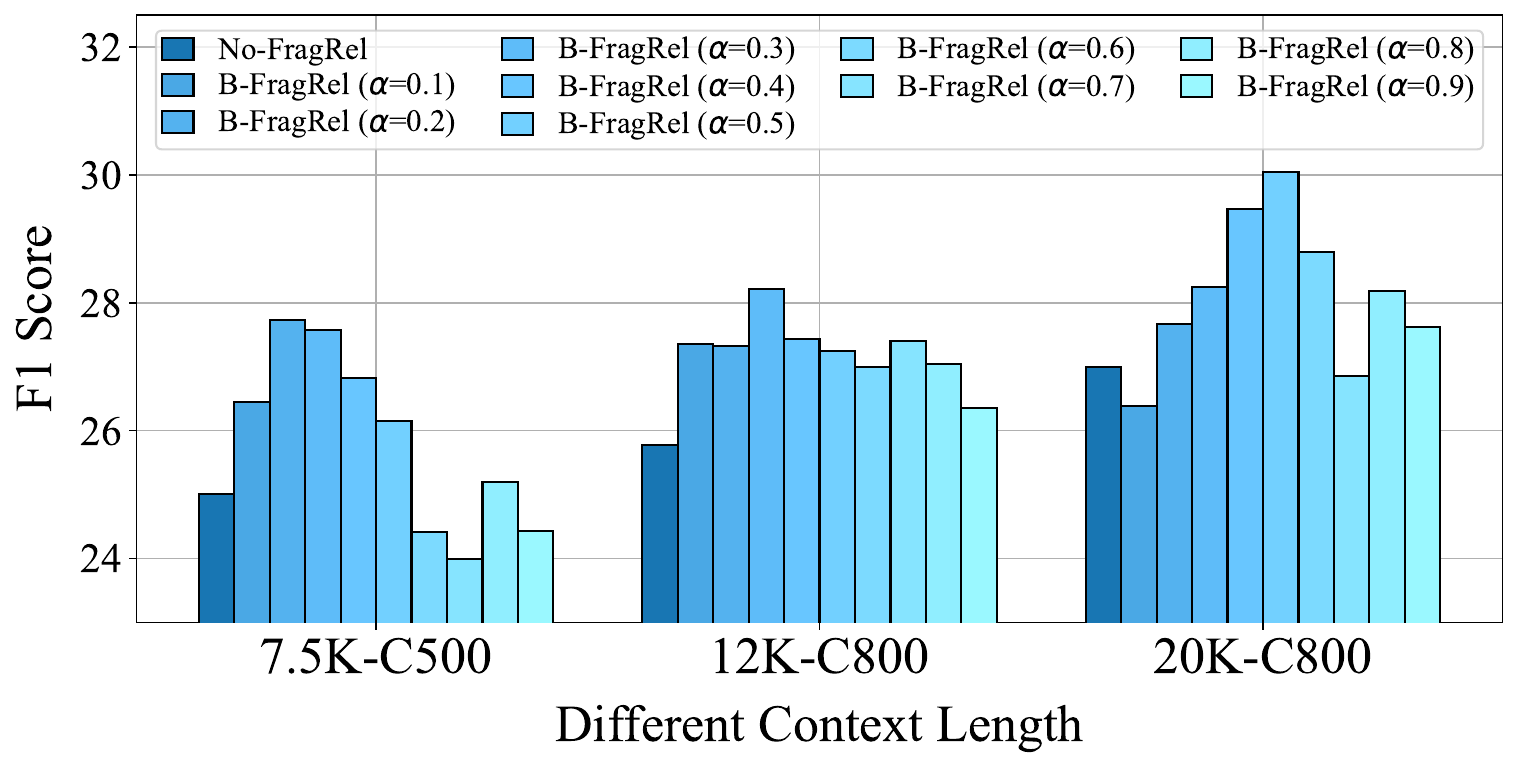}\label{fig:expres_ablation_rw}}
\vskip -0.1in
\caption{Performance improvement using different edge weight $w^{rel}$ and relation weight $\alpha$.}
\label{fig:expres_ablation}
\vskip -0.1in
\end{figure*}

We present the experimental results in Figure.~\ref{fig:expres_narra} and Figure.~\ref{fig:expres_ablation}.
The corresponding score value is reported in Table.~\ref{tab:apen_results_narra} and Table.~\ref{tab:apen_results_narra_ablation}, respectively.

\paragraph{TempMem \textit{vs.} HieraMem.}
According to Figure.~\ref{fig:expres_narra}, 
the Hierarchical Memory based LLMs (the columns) could outperform Temporary Memory based LLMs (the horizontal lines), especially on large context windows (more than 8K tokens).
This is consistent with the conclusion that retrieval augmentation could help improve long context LLM in the previous work~\citep{xu2023retrieval}.

\paragraph{Benefits of Fragment-level Relations.}
As shown in Figure.~\ref{fig:expres_narra}, across all base LLMs and context length, the structural relation augmented HieraMem-LLMs (noted as B-FragRel-*) consistently outperforms the counterpart HieraMem-LLMs (noted as No-FragRel-*).
The performance enhancement is especially noticeable under enough long  context length.
This indicates that the integration of fragment-level relations effectively alleviates the deficiencies of HieraMem-LLM in terms of external memory management.

\paragraph{Semantic Relation \textit{vs.} Structure Relation.}
Figure.~\ref{fig:expres_narra} compares semantic relation ("A-FragRel-*") and context structure relation ("B-FragRel-*") on fragment length 500. 
The semantic relation offers a slight enhancement under enough long context, while the context structure relation provides consistent and significant improvement across various context lengths. 
We posit this is due to that the embedding based retrieval has implicitly considered the semantic association between segments.

\paragraph{Different Relation Parameters.}
Figure.~\ref{fig:expres_ablation} presents the performance with varied relation parameters $w_{rel}$ and $\alpha$ on  different context lengths and fragment lengths.
Although the optimal values of relations parameters ($w_{rel}$ and $\alpha$) for different setups (including fragment length, context types, and context length constraint \etc) are difficult to ascertain,  most empirical values ($\alpha \in [0.2,0.5], w_{rel} \in [0.1,0.7]$) can lead to performance enhancement.

\subsection{Repository-level Code Generation}
\label{sec:exp_codegen}


\paragraph{Dataset.} 
We conduct the code generation experiment on RepoEval~\citep{zhang2023repocoder}, a benchmark constructed using the latest repositories source from GitHub. 
Specifically, two code completion tasks are considered: \textbf{(a)} \emph{Line Completion}: completing random code lines, \textbf{(b)} \emph{Api Invocation Completion}: completing random code lines that invoke in-repository apis.
Both tasks contain 1600 test samples across 8 repositories.

\paragraph{Metrics.}
Following previous works~\citep{zhang2023repocoder,DBLP:journals/corr/abs-2102-04664,lu2022reacc}, we evaluate the code generation performance using two metrics: \emph{Exact Match} (EM Score) and \emph{Edit Similarity} (ES Score).
EM score evaluates how many completions are exactly the same to real code.
ES score represents the similarity score between the generated and real code lines.

\paragraph{Baselines.}
We employ Codellama-34b~\citep{roziere2023code} with 4K context length as the base LLMs, and the same prompt formats as RepoCoder~\citep{zhang2023repocoder}. 
To extensively evaluate the integrated relations, 
we consider not only the single step Vanilla Retriever but also the iterative retrieval method RepoCoder~\citep{zhang2023repocoder}.
Noted that we report the result of oracle iterative retrieval, \ie the upper bound of performance during the iterative retrieval procedure.
Section.~\ref{sec:apen_framework_code} presents more implementation detail.

\paragraph{Relation Integration Details.}
In this experiment, we use the code structure relation shown in Equation.~\ref{equ:relation_instantiation_code_structure}
, and we set relation weight $\alpha=0.5$.


\paragraph{Performance Comparison.}
Table~\ref{tab:results_code_line},~\ref{tab:results_code_api} present the results of the line completion task and api invocation completion task, respectively.
On the two tasks, the integrated relation consistently improves the performance of both the single step retrieval inference framework and the iterative retrieval inference framework.
This empirically demonstrates that fragment-level relations are greatly helpful in long code scenarios.

\begin{table}[]
\center
\resizebox{\linewidth}{!}{
\begin{tabular}{ccc}
\toprule
\textbf{Method} & \textbf{EM Score} & \textbf{ES Score} \\
\midrule
\grayline \multicolumn{3}{c}{ \emph{ Single Step Retrieval } } \\
Vanilla Retriever & 46.31 & 66.26 \\
Vanilla Retriever + FragRel & \textbf{48.25}  & \textbf{67.05} \\
\midrule
\grayline \multicolumn{3}{c}{ \emph{ Iterative Retrieval } } \\
RepoCoder \citep{zhang2023repocoder} & 49.13  & 68.39        \\
RepoCoder + FragRel     &  \textbf{50.44} & \textbf{68.50} \\
\bottomrule
\end{tabular}
}
\caption{Performance evaluation on line completion task of RepoEval~\citep{zhang2023repocoder} using Codellama-34b~\citep{roziere2023code}.}
\label{tab:results_code_line}
\vskip -0.1in
\end{table}

\begin{table}[]
\center
\resizebox{\linewidth}{!}{
\begin{tabular}{ccc}
\toprule
\textbf{Method} & \textbf{EM Score} & \textbf{ES Score} \\
\midrule
\grayline \multicolumn{3}{c}{ \emph{ Single Step Retrieval } } \\
Vanilla Retriever & 40.00 & 66.32 \\
Vanilla Retriever + FragRel & \textbf{40.94}  & \textbf{66.39} \\
\midrule
\grayline \multicolumn{3}{c}{ \emph{ Iterative Retrieval } } \\
RepoCoder ~\citep{zhang2023repocoder} & 41.81  & 68.31        \\
RepoCoder + FragRel     &  \textbf{43.00} & \textbf{69.07} \\
\bottomrule
\end{tabular}
}
\caption{Performance evaluation on api invocation completion task of RepoEval~\citep{zhang2023repocoder} using Codellama-34b~\citep{roziere2023code}.}
\label{tab:results_code_api}
\vskip -0.2in
\end{table}

\begin{table*}
\center
\resizebox{1.0\linewidth}{!}{
\begin{tabular}{ccccc}
\toprule
\multirow{2}{*}{ \makecell{ \textbf{Method} } }  & \multicolumn{4}{c}{\textbf{Auto-rated Score by GPT4-4K (1-100)}} \\
\cmidrule(r){2-5} 
& \textbf{Retrospection} & \textbf{Continuation} & \textbf{Conjunction} & \textbf{Average} \\
\midrule
ChatGPT-2K &  52.11 & 55.33 & 48.22 & 51.89 \\
MPC-ChatGPT~\citep{lee2023prompted} &  53.00 & 61.22  & 49.33 & 54.52 \\
MemoryBank-ChatGPT~\citep{zhong2023memorybank} &  23.39 & 55.28  & 48.67 & 42.44 \\
MemoChat-ChatGPT~\citep{lu2023memochat} & 66.28 & 73.50  & 72.50 & 70.76 \\
\midrule
MemRetrieval-ChatGPT  &  81.17 & 74.56  & 69.39 & 75.04 \\
MemRetrieval-ChatGPT + FragRel  & \textbf{81.56} & \textbf{77.89}  & \textbf{82.17} & \textbf{80.54} \\	
\bottomrule
\end{tabular}
}
\caption{Performance comparison on MTBench+~\citep{lu2023memochat}.}
\label{tab:results_mtbench}
\vskip -0.2in
\end{table*}

\subsection{Memory-enhanced Chatbot}
\label{sec:exp_chatbot}

\paragraph{Dataset.} 
We perform the long-term chatting experiment on MTBench+~\citep{lu2023memochat}. 
Every chatting stream consists of 12 $\sim$ 15 turns dialogs, covering topics such as "STEM exams", and "literary writing".
At the end of every dialog, a challenging question is added by the experts. 
The questions could be classified into 3 categories:
\textbf{(a)} "Retrospection" requires the model to respond content mentioned previously, 
\textbf{(b)} "Continuation" requires the model to finish a further task about talked topics,
\textbf{(c)} "Conjunction" requires answering questions involving multiple topics existing in the dialog.
There are 54 test samples, 18 for every type.

\paragraph{Metrics.}
Following the origin benchmark~\citep{lu2023memochat}, we assess the generated response using LLM-as-a-judge method~\citep{zheng2023judging}, 
where GPT4 is instructed to check the faithfulness of the model response and produces a 1 $\sim$ 100 integer score.
We utilize exactly the same testing setup (including prompt format, GPT4 version, hyperparameters, \etc) as \citep{lu2023memochat}.

\paragraph{Baselines.}
We consider the ChatGPT-based baselines reported in MemoChat~\citep{lu2023memochat}, 
including various external memory enhanced frameworks~\citep{zhong2023memorybank,lee2023prompted}.
In addition, we introduce a dense retrieval augmentation baseline, named MemRetrieval-ChatGPT. 
Following previous work~\citep{lu2023memochat}, we constraint the temporary context length as 2K tokens.
Section.~\ref{sec:apen_framework_chatting} presents more implementation detail.

\paragraph{Relation Integration Details.}
Based on MemRetrieval-ChatGPT, we integrate the context structure relations defined in Equation.~\ref{equ:relation_instantiation_context_structure}.
We set $w_{rel}=0.8$ and $\alpha=0.5$ in our experiments.

\paragraph{Inference Expense Comparison.}
Approximately, MemoChat~\citep{lu2023memochat} consumes about 1M input tokens and 170K output tokens.
MemRetrieval (+FragRel) costs about 680K input tokens and 65K output tokens. 
The introduced dense retrieval framework relatively reduces about 32\% input tokens and 62\% output tokens cost.

\paragraph{Performance Comparison.}
Table~\ref{tab:results_mtbench} presents the experimental results.
Utilizing the same 2K temporary context length, the introduced MemRetrieval achieves comparable performance to MemoChat~\citep{lu2023memochat}.
Our fragment relations augmented methods consistently outperform all baselines, validating the effectiveness of using fragments relations on the long-term chatting task.

\section{Conclusion}
This work focuses on the challenge of isolated fragment processing in existing External Memory augmented Large Language Models (LLMs), proposing and formulating the fragment-level relations informed by intricate relations found within diverse long contexts. And we propose an efficacious method to integrate these fragment-level relations across distinct types of texts. Comprehensive experiments conducted over a range of long text processing tasks attest that the utilization of such fragment-level relations indeed enhances the performance of LLMs in various scenarios. We hope our findings will inspire further investigations into External Memory enhanced LLMs.

\section*{Limitation}

Despite notable enhancement on external memory augmented LLMs through integrating fragment-level relations, the current implementation still suffers the following limitations:

\noindent \textbf{1.} Manual relation definitions and empirical parameter selection: The relation definitions as outlined in Equations \ref{equ:relation_instantiation_semantic}, \ref{equ:relation_instantiation_context_structure}, \ref{equ:relation_instantiation_code_structure} are empirically defined, which limits their generalizability. Additionally, the optimal values for relation parameters ($w_{rel}$ and $\alpha$) vary based on text types, fragment lengths, and relation categories. We are, thus, limited to empirically selecting parameters that are good but not entirely optimal.
This problem may be resolved through automatic relations definition based on LLM-driven optimization strategies.

\noindent \textbf{2.} Inapplicability to arbitrary retrieval methods: The relation incorporation method introduced in Section~\ref{sec:relation_incorporation} only applies to retrieval methods that rely on fragment scores, thus neglecting other methods, such as generative retrieval.
This limitation is worth further investigation in future work.

\noindent \textbf{3.} Limited validation: Although we have substantiated the advantages of fragment-level relations through numerous long text processing tasks such as long story understanding, repository-level code completion, and memory-enhanced chatbot, we still recognize the necessity for validation on more extensive and complex tasks, such as academic material library understanding with near-infinite length of context, and multi-agents interactive tasks with more complex fragment-level relations in the external memory of LLMs.

\section*{Acknowledgment}

This work is supported by Fundamental Research Funds for the Central Universities (226-2023-00126). This work is also supported by the Major program of the National Natural Science Foundation of China (T2293720/T2293723).

\bibliography{reference}

\newpage
\appendix

\section{Code Relation Calculation Details}
\label{sec:apen_relation_code}

\paragraph{Code Repository Graph $\mathcal{G}$ Construction.}
The code repository graph is created with the following steps:
(1) For a specific code repository, 
we first construct the code syntax tree for every code source file using tree-sitter~\footnote{https://tree-sitter.github.io}  (edge weight is set as 0.5).
(2) In addition, we construct the file directory tree where every node corresponds to a file or directory in the repository (edge weight is set as 0.3).
(3) Next, we connect the root node of every code syntax tree with its counterpart file node on the file directory tree (edge weight is set as 1.0), obtaining the code repository parsing tree.
(4) Finally, we create node connections (edge weight is set as 0.8) between every function invoking node and its corresponding function definition node (including both in-file invoking and cross-file invoking).

\paragraph{Distance calculation.}
On code repository graph $\mathcal{G}$, the edge weight is set according to the edge types (a connection between files, a connection between in-file code syntax node, \etc), and constrained in $[0,1]$.

The length of a path through multiple nodes is defined as the product of the weights of all edges on the path, thus a longer path will have a smaller weight. 
The relation strength $\text{Dis}(g_k^{c_i},g_l^{c_i})$ of the relationship between two nodes is defined as the weight of maximum weight path between the two nodes $g_k^{c_i}$ and $g_l^{c_i}$.
We calculate $\text{Dis}(\cdot,\cdot)$ via the Dijkstra algorithm in our experiment.

\section{Framework Implementation Details}
\label{sec:apen_framework_detail}

\subsection{Long Story Understanding Details}
\label{sec:apen_framework_narrativeqa}

\paragraph{Prompt Template.}
We provide the prompt template used in the experiment of long story understanding in Prompt. 1.

\paragraph{Retrieval.}
We take same fragment splitting and embedding calculation method as LongBench~\citep{bai2023longbench}.
Contriever~\citep{izacard2021contriever} is used as the text embedding model.
We experiment different fragment length (500 and 800 words) for different context length limitations.

\paragraph{LLM Inference.}
In the experiment, same context content are provided for different LLMs.
The detail prompt template slightly varies in different LLMs according to their training template.
For example, the prompt is covered with "[INST]" and "[/INST]" for Llama2.
The temperature is set as $1.0$ during inference.

\subsection{Code Completion Framework Details}
\label{sec:apen_framework_code}

\paragraph{Prompt Template.}
We provide the prompt template used in the experiment of repository-level code generation in Prompt. 2.


\paragraph{Retrieval.}
Every fragment consists of exactly $S_w$ lines of code and adjacent fragments have $S_s$ overlapped lines of code.
We set $S_w=20$ and $S_s=10$ same as \citet{zhang2023repocoder}.
Following RepoCoder~\citep{zhang2023repocoder}, the BM25~\citep{10.1561/1500000019} is used for retrieval.
The fragments with top~10 similarity scores to the completed code context are taken as retrieved results every time.

\paragraph{LLM Inference.}
We employ Codellama-34b~\citep{roziere2023code} with 4K context length as the base LLMs in our experiment, and exactly the same prompt template as used in RepoCoder~\citep{zhang2023repocoder}.

\subsection{Memory-enhanced Chatbot Details}
\label{sec:apen_framework_chatting}

\paragraph{Prompt Template.}
We provide the prompt template used in the experiment of chatbot in Prompt. 3 (for inference) and Prompt. 4 (for evaluation).


\paragraph{Retrieval.}
In MemRetrieval-ChatGPT (+FragRel), every fragment consists of exactly one turn of the dialog.
The text embedding is calculated using the text-embedding-ada-002 model.
Every time we load related historical fragments with top 8 similarity scores to the recent dialog (last turn of dialog and latest user prompt).
Additionally, the retrieved fragments are reordered according to their chatting time.

\paragraph{LLM Inference.}
Same as MemoChat~\citep{lu2023memochat}, we use the GPT-3.5-Turbo as the base LLM for inference, and the context length is constrained to 2K tokens.
In the initial rounds of conversation, all historical records are directly inputted into the model. 
Once the length of the historical record exceeds 1K tokens or the conversation rounds surpass 10, the historical chat content will be stored in external memory. 
Subsequently, each conversation will load relevant fragments to the recent chatting context from the external memory.

\begin{figure*}
\begin{lstlisting}[breaklines=true,caption={Prompt template in the experiment of long story understanding.}]
You are given a story, which can be either a novel or a movie script, and a question. Answer the question asconcisely as you can, using a single phrase if possible. Do not provide any explanation.

Story: {context}

Now, answer the question based on the story asconcisely as you can, using a single phrase if possible. Do not provide any explanation.

Question: {input}

Answer:
\end{lstlisting}
\vskip -0.2in
\end{figure*}
\begin{figure*}
\begin{lstlisting}[breaklines=true,caption={Prompt template in the experiment of code generation.}]
# Here are some relevant code fragments from other files of the repo:
# -----------------
# the below code fragment can be found in:
# ***.py
# -----------------
# [RETRIEVED_CODE_CONTENT]
# -----------------
# ***.py
# -----------------
# [RETRIEVED_CODE_CONTENT]
# -----------------
# ***.py
# -----------------
# [RETRIEVED_CODE_CONTENT]
# -----------------
...

[CURRENT_FILE_CONTENT]
...
\end{lstlisting}
\vskip -0.2in
\end{figure*}
\begin{figure*}
\begin{lstlisting}[breaklines=true,caption={System prompt used in the experiment of chatbot.}]
You are an intelligent dialog bot. You will be shown Related Evidences supporting for User Input, and Recent Dialogs between user and you. Please read, memorize, and understand given materials, then generate one concise, coherent and helpful response.
\end{lstlisting}
\vskip -0.2in
\end{figure*}
\begin{figure*}
\begin{lstlisting}[breaklines=true,caption={Prompt template for evaluating the generated results with GPT-4.}]
You are an impartial judge. You will be shown Related Conversation History, User Question and Bot Response.

```
Related Conversation History
[RCH_0]
```

```
User Question
[UQ_1]
```

```
Bot Response
[BR_2]
```

Please evaluate whether Bot Response is faithful to the content of Related Conversation History to answer User Question. Begin your evaluation by providing a short explanation, then you must rate Bot Response on an integer rating of 1 to 100 by strictly following this format: [[rating]].
\end{lstlisting}
\vskip -0.2in
\end{figure*}

\begin{table}[]
\center
\resizebox{\linewidth}{!}{
\begin{tabular}{ccc}
\toprule
\textbf{Context Length} & \textbf{Method} & \textbf{F1 Score} \\
\midrule
\grayline \multicolumn{3}{c}{\emph{Temporary Memory based LLMs}} \\
\midrule
\multicolumn{2}{c}{GPT-3.5-Turbo-16k} &  23.60 \\
\multicolumn{2}{c}{ChatGLM3-6B-32k}  &  26.00 \\
\multicolumn{2}{c}{LongChat-v1.5-7B-32k}  & 16.90 \\
\multicolumn{2}{c}{XGen-7B-8k}  &  18.00 \\
\multicolumn{2}{c}{InternLM-7B-8k}  &  12.10 \\
\multicolumn{2}{c}{Vicuna-v1.5-7B-16k}  &  19.40 \\
\multicolumn{2}{c}{Llama2-7B-chat-4k}  &  18.74 \\
\midrule
\grayline \multicolumn{3}{c}{\emph{Hierarchical Memory based LLMs}} \\
\midrule
1K &  No-FragRel-ChatGLM-C500 &       18.81 \\
1K &  A-FragRel-ChatGLM-C500 &        18.32 \\
1K &  B-FragRel-ChatGLM-C500 &        19.71 \\
1K &  No-FragRel-ChatGPT-C500 &       17.19 \\
1K &  B-FragRel-ChatGPT-C500 &        17.61 \\
1K &  No-FragRel-Llama2-C500 &        15.75 \\
1K &  B-FragRel-Llama2-C500 &         17.30 \\
\midrule
2K &  No-FragRel-ChatGLM-C500 &       20.47 \\
2K &  A-FragRel-ChatGLM-C500 &        20.21 \\
2K &  B-FragRel-ChatGLM-C500 &        20.63 \\
2K &  No-FragRel-ChatGPT-C500 &       20.15 \\
2K &  B-FragRel-ChatGPT-C500 &        20.20 \\
2K &  No-FragRel-Llama2-C500 &        17.44 \\
2K &  B-FragRel-Llama2-C500 &         18.02 \\
\midrule
4K &  No-FragRel-ChatGLM-C500 &       22.32 \\
4K &  A-FragRel-ChatGLM-C500 &        22.90 \\
4K &  B-FragRel-ChatGLM-C500 &        24.43 \\
4K &  No-FragRel-ChatGLM-C800 &       26.06 \\
4K &  B-FragRel-ChatGLM-C800 &        26.26 \\
4K &  No-FragRel-ChatGPT-C500 &       25.07 \\
4K &  B-FragRel-ChatGPT-C500 &        25.11 \\
4K &  No-FragRel-ChatGPT-C800 &       22.25 \\
4K &  B-FragRel-ChatGPT-C800 &        24.50 \\
4K &  No-FragRel-Llama2-C500 &        18.22 \\
4K &  B-FragRel-Llama2-C500 &         20.32 \\
\midrule
8K &  No-FragRel-ChatGLM-C500 &       27.44 \\
8K &  A-FragRel-ChatGLM-C500 &        26.21 \\
8K &  B-FragRel-ChatGLM-C500 &        28.07 \\
8K &  No-FragRel-ChatGLM-C800 &       27.38 \\
8K &  B-FragRel-ChatGLM-C800 &        28.48 \\
8K &  No-FragRel-ChatGPT-C500 &       24.86 \\
8K &  B-FragRel-ChatGPT-C500 &        28.25 \\
8K &  No-FragRel-ChatGPT-C800 &       24.06 \\
8K &  B-FragRel-ChatGPT-C800 &        26.04 \\
\midrule
12K &         No-FragRel-ChatGLM-C500 &       28.20 \\
12K &         A-FragRel-ChatGLM-C500 &        27.56 \\
12K &         B-FragRel-ChatGLM-C500 &        29.55 \\
12K &         No-FragRel-ChatGLM-C800 &       25.78 \\
12K &         B-FragRel-ChatGLM-C800 &        27.57 \\
\midrule
20K &         No-FragRel-ChatGLM-C500 &       26.71 \\
20K &         A-FragRel-ChatGLM-C500 &        27.08 \\
20K &         B-FragRel-ChatGLM-C500 &        28.79 \\
20K &         No-FragRel-ChatGLM-C800 &       26.99 \\
20K &         B-FragRel-ChatGLM-C800 &        27.93 \\
\midrule
28K &         No-FragRel-ChatGLM-C500 &       27.51 \\
28K &         A-FragRel-ChatGLM-C500 &        28.05 \\
28K &         B-FragRel-ChatGLM-C500 &        30.41 \\
28K &         No-FragRel-ChatGLM-C800 &       27.28 \\
28K &         B-FragRel-ChatGLM-C800 &        29.43 \\
\bottomrule
\end{tabular}
}
\caption{The quantity value of experimental results on NarrativeQA, \ie the results shown in Figure.~\ref{fig:expres_narra}}
\label{tab:apen_results_narra}
\end{table}

\begin{table}[]
\center
\resizebox{0.90\linewidth}{!}{
\begin{tabular}{ccccc}
\toprule
\multirow{2}{*}{\textbf{Fragments}} & \multicolumn{2}{c}{\textbf{Ablation on} $\alpha$} & \multicolumn{2}{c}{\textbf{Ablation on} $w_{rel}$}  \\
 & \textbf{$\alpha$} & F1 Score & \textbf{$w_{rel}$} &  F1 Score \\
\midrule
No-FragRel & - & 25.01 & - & 25.01 \\
7.5K-C500 & 0.1 &  26.45  & 0.1  & 26.15 \\
7.5K-C500 & 0.2 &  27.73  & 0.2  & 25.81  \\
7.5K-C500 & 0.3 &  27.57  & 0.3  & 26.75  \\
7.5K-C500 & 0.4 &  26.82  & 0.4  & 26.52 \\
7.5K-C500 & 0.5 &  26.15  & 0.5  & 26.67  \\
7.5K-C500 & 0.6 &  24.42  & 0.6  & 26.78  \\
7.5K-C500 & 0.7 &  23.99  & 0.7  & 26.38  \\
7.5K-C500 & 0.8 &  25.20  & 0.8  & 25.24  \\
7.5K-C500 & 0.9 &  24.43  & 0.9  & 25.34  \\
\midrule
No-FragRel & - & 25.78 & - & 25.78 \\
12K-C800 & 0.1 &  27.35  & 0.1  & 27.25 \\
12K-C800 & 0.2 &  27.32  & 0.2  & 26.83  \\
12K-C800 & 0.3 &  28.22  & 0.3  & 28.83  \\
12K-C800 & 0.4 &  27.43  & 0.4  & 28.85 \\
12K-C800 & 0.5 &  27.25  & 0.5  & 29.90  \\
12K-C800 & 0.6 &  27.00  & 0.6  & 28.48  \\
12K-C800 & 0.7 &  27.4   & 0.7  & 27.57  \\
12K-C800 & 0.8 &  27.04  & 0.8  & 27.15  \\
12K-C800 & 0.9 &  26.35  & 0.9  & 27.63  \\
\midrule
No-FragRel & - & 26.99 & - & 26.99 \\
20K-C800 & 0.1 & 26.39 & 0.1  & 30.05 \\
20K-C800 & 0.2 & 27.66 & 0.2  & 29.46  \\
20K-C800 & 0.3 & 28.24 & 0.3  & 29.30  \\
20K-C800 & 0.4 & 29.46 & 0.4  & 28.30 \\
20K-C800 & 0.5 & 30.05 & 0.5  & 27.90  \\
20K-C800 & 0.6 & 28.80 & 0.6  & 28.41  \\
20K-C800 & 0.7 & 26.86 & 0.7  & 27.71  \\
20K-C800 & 0.8 & 28.19 & 0.8  & 27.60  \\
20K-C800 & 0.9 & 27.62 & 0.9  & 27.93  \\
\bottomrule
\end{tabular}
}
\caption{The quantity value of ablation study results on NarrativeQA, \ie the results shown in Figure.~\ref{fig:expres_ablation}.}
\label{tab:apen_results_narra_ablation}
\end{table}

\end{document}